\documentclass[10pt,twocolumn,letterpaper]{article}

\usepackage{wacv}
\usepackage{times}
\usepackage{epsfig}
\usepackage{graphicx}
\usepackage{amsmath}
\usepackage{amssymb}
\usepackage{subcaption}
\usepackage{multirow}
\usepackage{hhline}
\usepackage{hyperref}
\wacvfinalcopy % *** Uncomment this line for the final submission

\def\wacvPaperID{87} % *** Enter the wacv Paper ID here

\ifwacvfinal\fi
\begin{document}

%%%%%%%%% TITLE
\title{B-CNN: Branch Convolutional Neural Network for Hierarchical Classification}

% Authors at the same institution
%\author{First Author \hspace{2cm} Second Author \\
% Authors at different institutions
\author{Xinqi Zhu \\
Dept. of Computer Science and Engineering\\
Shanghai Jiao Tong University\\
{\tt\small zhuxinqimac@sjtu.edu.cn}
\and
Michael Bain \\
School of Computer Science and Engineering\\
University of New South Wales\\
{\tt\small m.bain@unsw.edu.au}
}

\maketitle
\ifwacvfinal\thispagestyle{empty}\fi

%%%%%%%%% ABSTRACT
\begin{abstract}
Convolutional Neural Network (CNN) image classifiers are traditionally designed to have sequential convolutional layers with a single output layer. This is based on the assumption that all target classes should be treated equally and exclusively. However, some classes can be more difficult to distinguish than others, and classes may be organized in a hierarchy of categories. At the same time, a CNN is designed to learn internal representations that abstract from the input data based on its hierarchical layered structure. So it is natural to ask if an \emph{inverse} of this idea can be applied to learn a model that can predict over a classification hierarchy using multiple output layers in \emph{decreasing} order of class abstraction. In this paper, we introduce a variant of the traditional CNN model named the Branch Convolutional Neural Network (B-CNN). A B-CNN model outputs multiple predictions ordered from coarse to fine along the concatenated convolutional layers corresponding to the hierarchical structure of the target classes, which can be regarded as a form of prior knowledge on the output. To learn with B-CNNs a novel training strategy, named the Branch Training strategy (BT-strategy), is introduced which balances the strictness of the prior with the freedom to adjust parameters on the output layers to minimize the loss. In this way we show that CNN based models can be forced to learn successively coarse to fine concepts in the internal layers at the output stage, and that hierarchical prior knowledge can be adopted to boost CNN models' classification performance. Our models are evaluated to show that the B-CNN extensions improve over the corresponding baseline CNN on the benchmark datasets MNIST, CIFAR-10 and CIFAR-100.
\end{abstract}

%%%%%%%%% BODY TEXT
\section{Introduction}
\label{introduction}
The traditional CNN based classification models are designed to be sequential and output the only prediction at the top of the models without any branching of the network at the outputs. This is because these models assume that all classes are equally difficult to distinguish and treat all of them exclusively. But in fact, the property of general-to-specific category ordering often exists between classes, \eg, \emph{cat} and \emph{dog} can usually be grouped as \emph{pet} while \emph{chair} and \emph{bed} are \emph{furniture}, and it is often easier to tell a cat apart from a bed than from a dog. This property indicates that classification can be done in a hierarchical way instead of treating all classes as arranged in a ``flat'' structure. When doing hierarchical classification, a classifier first knows an apple should be in the coarse category of \emph{fruit}, then it can be classified at the finer level as an \emph{apple}. One benefit of hierarchical classification is that the error can be restricted to a subcategory, which also means it should be more informative than flat classification. For example, a classifier may confuse an apple with an orange, but knows it should at least be \emph{fruit} so won't confuse it with, say, a red snooker ball.

%Though traditional CNN models can perform really well when doing flat classification tasks, they are lack of the ability to control the errors in a narrow concept which should be more easily understandable by human. For example, it is usually more acceptable if an apple is incorrectly classified as an orange instead of a Snooker red ball as you still get the information that apple should be a breed of fruit. Traditional CNNs also require a huge number of training examples. However some problems do not have big enough training set to support each class. In this situation, shared features between classes become important as we can borrow knowledge from relevant classes.

CNN based models are naturally hierarchical. As Zeiler and Fergus present in \cite{Visualizing}, lower layers in CNN usually capture the low level features of an image such as basic shapes while higher layers are likely to extract high level features such as the face of a dog. As a consequence, a possible way to embed a hierarchy of classes into a CNN model is to output multiple predictions along the CNN layers as the data flow through, from coarse to fine. In this case, lower layers output coarser predictions while higher layers output finer predictions. Unlike traditional CNN models which do not capture the complexity of semantic labels in the real world, hierarchical classification models can do predictions in a more interpretable way and even boost the final classification as the hierarchical prior is a good guide to the classifier.

%This concept-shared classification can be achieved by adopting the hierarchical relations between labels. Unlike traditional CNN models which do not capture the complexity of semantic labels in the real world, hierarchical relations can constrain the prediction results in a sub category and sometimes even boost the classification performance. For example, when dealing with three classes \emph{cat}, \emph{dog} and \emph{airplane}, it is easy to tell \emph{airplane} from the other two classes with low level features such as basic shape. If we can separate \emph{airplane} from \emph{cat} and \emph{dog} early, then the classifier can focus on the details which can tell \emph{cat} and \emph{dog} apart without considering any distraction from \emph{airplane}.

\begin{figure*}
    \begin{center}
    \begin{subfigure}[t]{0.7\textwidth}
        \centering
        \includegraphics[width=\textwidth]{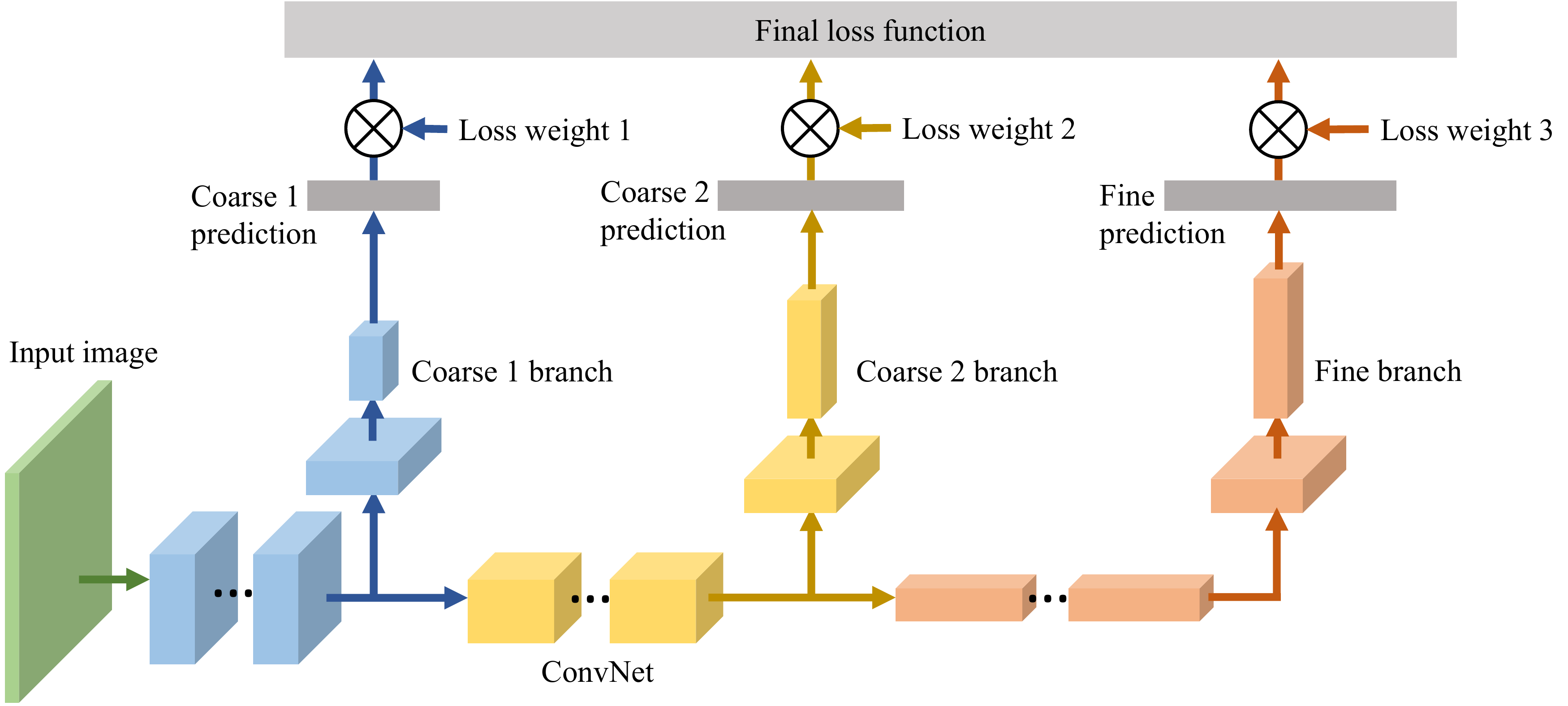}
        \caption{}
        \label{img:arch}
    \end{subfigure}%
    ~ 
    \begin{subfigure}[t]{0.3\textwidth}
        \centering
        \includegraphics[width=\textwidth]{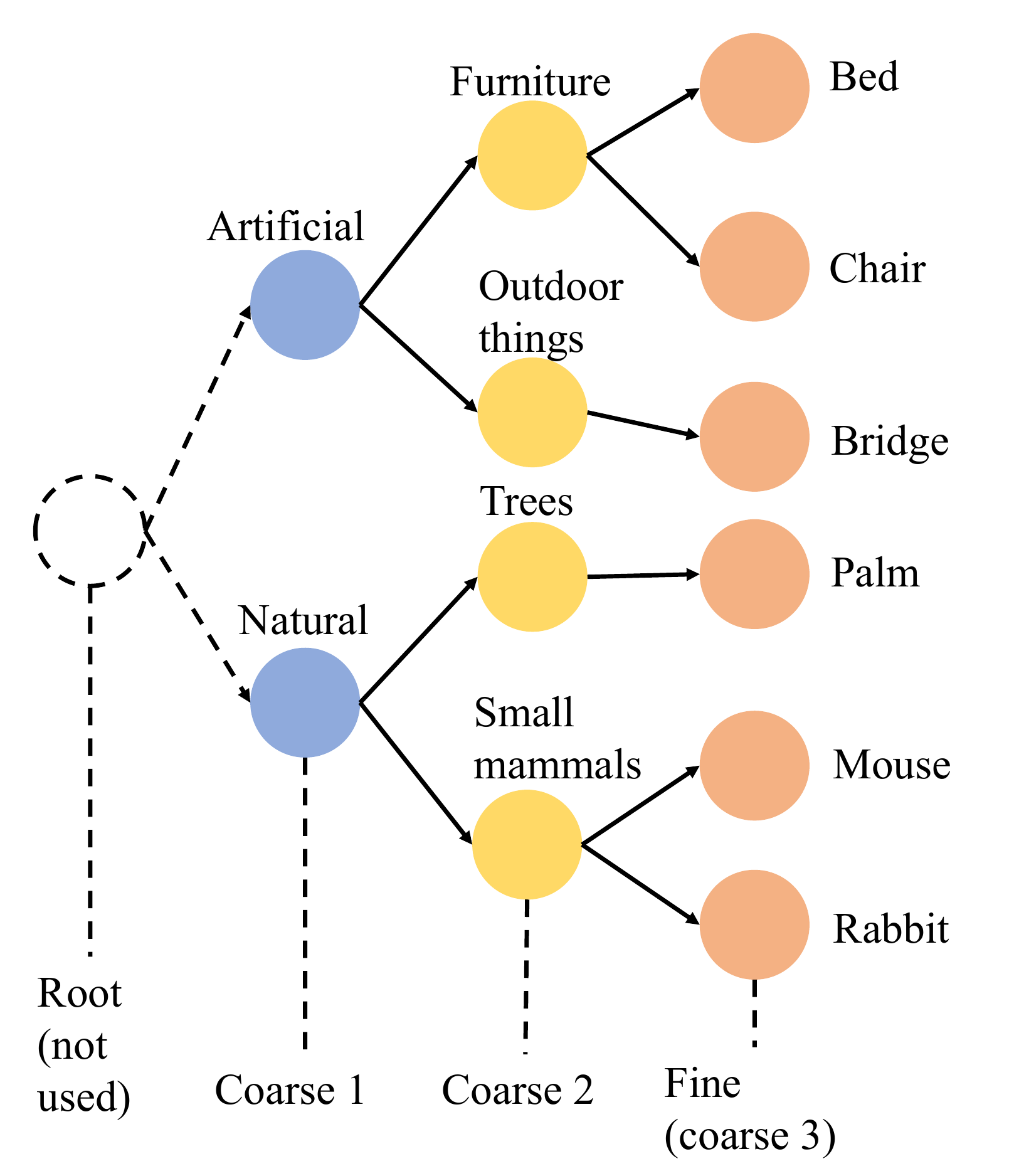}
        \caption{}
        \label{img:tree}
    \end{subfigure}
%     \fbox{\rule{0pt}{2in} \rule{.9\linewidth}{0pt}}
    \end{center}
    \caption{(a) Architecture of Branch Convolutional Neural Network (B-CNN). The network at the bottom can be an arbitrary ConvNet. There can be multiple branch networks and each of them outputs a coarse prediction. The final loss function is a weighted summation of all coarse losses. (b) A sample hierarchical label tree where classes are taken from CIFAR-100 dataset.}
    \label{img:img}
\end{figure*}

In this paper, we introduce a special CNN based model integrated with the prior knowledge of hierarchical category relations (\url{https://github.com/zhuxinqimac/B-CNN}). We name it \emph{Branch Convolutional Neural Network} (B-CNN) as it contains several branch networks along the main convolution workflow to do predictions hierarchically. The architecture of B-CNN is shown in Figure \ref{img:arch}. Our B-CNN model is inspired by the idea from \cite{Visualizing} that each layer of a CNN contains the hierarchical nature of the features in the network. We exploit this property of CNN to combine it with the prior knowledge of class hierarchical structure to enforce the network to learn human understandable concepts in different layers.

Beside B-CNN, we also propose a novel training strategy which is tailored to our B-CNN models named \emph{Branch Training strategy} (BT-strategy). When training a B-CNN model with BT-strategy, lower level parameters are activated and trained earlier than the higher level parameters. This idea is inspired by the vanishing gradient problem \cite{vanishinggradient} and the step-by-step training method adopted by Simonyan and Zissermanto to train a deep ConvNet in \cite{vgg}. BT-strategy can prevent the impact of vanishing gradient problem to an extent and boost the performance of a B-CNN model.

%Our B-CNN model is inspired by the idea from \cite{Visualizing} that each layer of a CNN contains the hierarchical nature of the features in the network. We exploit this property of CNN to combine it with the prior knowledge of class hierarchical structure to enforce the network to learn human understandable concepts in different layers. This not only ensures the classifier to be trained in a feature-shared way, but also increases the interpretability of the network significantly.

This paper's major contributions are summarized below. First, we introduce a new CNN based model for hierarchical classification. Second, our work presents the possibility of embedding the semantic structure of target classes into a CNN model to demonstrate the interpretability of convolutional neural network. Third, we propose a novel training strategy tailored to B-CNN which can boost the classification accuracy. We validate our model and training strategy on MNIST, CIFAR-10 and CIFAR-100 datasets, showing significant empirical benefits.

%-------------------------------------------------------------------------

\section{Related Work}
\label{related_work}
CNN based models have been successfully exploited in many computer vision tasks and the most successful one is image classification \cite{alexNet}. There are various techniques introduced to enhance CNN's performance in the past literature. Simonyan and Zisserman show us that increasing the depth of a CNN model \cite{vgg} within a boundary is likely to boost the classifier's accuracy. In \cite{Visualizing} and \cite{OverFeat}, a large number of filters has been adopted. Recently, there have been a vast amount of work to enhance the components of CNN such as novel types of activation functions \cite{ELU, maxout}, the replacement of linear filter patches \cite{NIN}, pooling operation \cite{generalizingPooling, strivingForSimplicity, StochPooling} and initialization strategies \cite{init}. These attempts mainly focus on the potential weakness inside the existent CNN models and try to find a replacement or propose any strategy to improve it. Different from them, our B-CNN model is taking existent CNNs as building blocks, exploiting natural hierarchical property of CNN \cite{Visualizing}, and boosting performance with a tailored training strategy.
% to do a more informative and interpretable classification.

Exploiting hierarchical structure of object category has a long history \cite{hierarchySurvey}. In \cite{VisualConceptLearning, ExploitingObjectHierarchy}, hierarchy in classes has been used to combine different models for better performance. The hierarchy of classes can either be predefined by human \cite{SemanticHierarchies2007, LearningHierarchicalSimilarityMetrics, ExploitingObjectHierarchy}, or constructed automatically by top-down and bottom-up approaches \cite{HierarchicalImageAnnotation, BuildingAndUsing, ConstructingCategoryHierarchies, LearningToShare}.

Combining tree structure prior with CNN models has drawn interest recently. Srivastava \etal \cite{DiscriminativeTransfer} propose a method which benefits from CNN and tree-based prior when the training set is very small. They have shown that label tree prior can be used to transfer knowledge between classes and boost the performance when training examples are insufficient. But their method does not exploit the hierarchical nature within the layers which remains the whole CNN model a black box. Jia \etal \cite{LargeScaleObject} introduce a graph representation to capture the hierarchical and exclusive semantics between labels. They define a joint distribution of an assignment of all labels as a Conditional Random Field and use it to replace the traditional classifiers such as softmax on the top of a deep neural network. However their work still uses the network as a feature extractor and there is no attempt to exploit the hierarchical nature of CNN itself.

Zhicheng \etal \cite{HDCNN} have made a significant contribution in discovering the possibility of combing the label tree and the hierarchical nature of CNN. In their work, classes are grouped as coarse categories and their HD-CNN model separates easy categories using a shared coarse classifier while distinguishing difficult classes using fine category classifiers. This model confirmed that the hierarchical property in CNN can be exploited. A potential problem of HD-CNN is that it requires the coarse and fine category components to be pretrained and followed by a fine-tune procedure which is quite time-consuming. HD-CNN also only uses one coarse category which is not scalable because there may be far more than one levels of coarse categories in a hierarchical label tree.

% Learning DNNs with Probabilistic Graphical Models (MRFs, CRFs, etc.) as
% output (Chen_etal_2015b.pdf MRFs; CRF-CN.pdf)
% Structured Prediction Energy Networks (Belanger_McCallum_2016.pdf)
% Jain_Structural-RNN_Deep_Learning_CVPR_2016_paper
More recently work has combined deep learning with probabilistic graphical
models such as MRFs, CRFs, etc. (\eg~\cite{Chen:etal:j:2016}), but
this either requires a two-stage training process, or new methods of
training the combined model, which is not as straightforward as training
in a B-CNN. 
An end-to-end approach is the Structured Prediction Energy Networks
of~\cite{Bela:McCa:p:2016} but this is not based on CNNs.

The vanishing gradient problem \cite{vanishinggradient} has existed in neural networks for a long time and various methods have been proposed to handle it. Rectified linear units (ReLU) have been successfully adopted in deep neural networks \cite{relu} as the activation function. LSTM models use forget gates to solve the vanishing gradient issue in RNNs \cite{LearningToForget, lstm}. Recently, residual neural networks (ResNets) eliminate this problem by adding identity shortcut connections between three layer chunks in ConvNets \cite{ResNet}. Our BT-strategy relieves this issue in B-CNNs by shifting the loss weights of different coarse level outputs, ultimately making it converge to a traditional CNN classifier.

%Vanishing gradient problem \cite{vanishinggradient} exists in neural network for a long time and there are various of methods proposed to handle it. Rectified linear units (ReLU) are successfully adopted in deep neural networks \cite{relu} as the activation function. LSTM models use forget gates to solve the vanishing gradient issue in RNNs \cite{LearningToForget, lstm}. Recently, residual neural networks (ResNets) eliminate this problem by adding identity shortcut connections between three layer chunks in ConvNets \cite{ResNet}. Our BT-strategy relieves this issue in B-CNNs by shifting the loss weights of different coarse level outputs and making it converge to a traditional CNN classifier at last.

\section{Model Description}
\label{architecture}

\subsection{Branch Convolutional Neural Network}
The overview architecture of \emph{Branch Convolutional Neural Network} (B-CNN) is shown in Figure \ref{img:arch} and a corresponding label tree is in Figure \ref{img:tree}. In the label tree, fine labels are target classes and are always provided by the classification task. They are presented as leaves and clustered into coarse categories which can be manually constructed or generated by unsupervised methods. More general categories can be used, \eg, coarse 1 level in Figure \ref{img:tree} is a super-level of coarse 2. In order not to cause ambiguity, we use \emph{level} to refer to the different layers in the label tree, \emph{layer} as the layers in the neural network and \emph{branch} as the branch output nets of B-CNN.

A B-CNN model uses existent CNN components as building blocks to construct a network with internal output branches. The network shown at the bottom in Figure \ref{img:arch} is a traditional convolutional neural network. It can be an arbitrary ConvNet with multiple layers. The middle part in Figure \ref{img:arch} shows the output branch networks of a B-CNN. Each branch net produces a prediction on the corresponding level in the label tree (Figure \ref{img:tree}, shown in same color). On the top of each branch, fully connected layers and a softmax layer are used to produce the output in one-hot representation. Branch nets can consist of ConvNets and fully connected neural networks. But for simplicity, in our experiments, we only use fully connected neural networks as our branch nets.

When doing classification, a B-CNN model outputs as many predictions as the levels the corresponding label tree has. For example, considering the label tree shown in Figure \ref{img:tree}, an image of a mouse will contain a hierarchical label of [\emph{natural, small mammals, mouse}]. When the image is fed into B-CNN, the network will output three corresponding predictions as the data flow through and each level's loss will contribute to the final loss function (introduced in \ref{loss_function}) base on the loss weights distribution (introduced in \ref{loss_weight}).

\subsection{Loss Function}
\label{loss_function}

The loss function of B-CNN is a weighted summation of all coarse and find prediction losses. The loss function is defined in (\ref{eq:loss}):

\begin{equation} \label{eq:loss}
L_{i} = \sum_{k=1}^{K}-A_{k}\log{\bigg(\frac{e^{f_{y_{i}}^{k}}}{\sum_{j}e^{f_{j}^{k}}}\bigg)}
\end{equation}
where $i$ denotes the $i^{th}$ sample in the mini-batch. $K$ is the number of coarse levels in the label tree and $A_{k}$ is the loss weight (introduced in \ref{loss_weight}) of $k^{th}$ level contributing to the loss function. The term $-\log{\bigg(\frac{e^{f_{y_{i}}^{k}}}{\sum_{j}e^{f_{j}^{k}}}\bigg)}$ is the cross-entropy loss of the $i^{th}$ sample on the $k^{th}$ level in the label tree and we use $f_{j}$ to denote the $j^{th}$ element in  the vector $f$ of class scores, outputted by the last layer of the model.

The loss function takes all levels' loss into account to make sure the structure prior can play a role of internal guide to the whole model and make it easier to flow the gradients back to the shallow layers.

\subsection{Loss Weight}
\label{loss_weight}

Value $A_{k}$ in (\ref{eq:loss}) is the loss weight of each branch network in B-CNN. This value defines how much contribution a level makes to the final loss function. Although only relative value matters how the loss function works, we use a standard representation in our experiments that each $A_{k}$ should be a value between 0 and 1 (both included) and their summation should be 1, \eg, we use loss weights [\emph{0.1, 0.1, 0.8}] for a three level label tree instead of [\emph{1, 1, 8}]. 
%Weight $A_{k}$ is defined in (\ref{eq:weight}):
%\begin{equation} \label{eq:weight}
%A_{k} = A_{K} \times 10^{\sqrt{K-k}}
%\end{equation}
%for simplicity, here we refer the fine level as coarse $K$ level and its weight $A_{K}$ is fixed to 1. Equation (\ref{eq:weight}) is a rough guide to select $A_{k}$s and can be replaced by manually selected values for better experimental results. Basically, $A_{i} > A_{i+1}$ should hold as when general categories are correctly classified, specific labels are more likely to be correct.

This definition of loss weight also makes the B-CNN a super model of traditional CNN. For instance, for a three-branch B-CNN (corresponding to a three level label tree), when loss weights are fixed to [\emph{0, 0, 1}], the B-CNN model converges to a traditional CNN model with only the last output branch trainable. With loss weights of [\emph{1, 0, 0}], the B-CNN will only activate the former part of the whole network remaining the two higher levels untrained. The distribution of loss weights also indicates the importance of each level, \eg, loss weights of [\emph{0.98, 0.01, 0.01}] mean the model values the low level feature extraction but also want to train a little bit of the deep layers, and in our experiments, we usually use this assignment as the initialization of our loss weights.

\subsection{Branch Training Strategy}
Our \emph{Branch Training strategy} (BT-strategy) exploits the potential of loss weights distribution to achieve an end-to-end training procedure with low impact of vanishing gradient problem \cite{vanishinggradient}. BT-strategy modifies the loss weights distribution while training a B-CNN model, \eg, for a two level classification, the initial loss weights can be assigned as [\emph{0.9, 0.1}], and then they can be changed to [\emph{0.2, 0.8}] after 50 epochs.

The greatest value in a loss weights assignment can be regarded as a ``focus'', \eg, 0.5 is the focus of [\emph{0.2, 0.3, 0.5}]. A ``focus'' is not necessary in a distribution as all levels can be equally important. However, in our implementation, we usually set a ``focus'' to explicitly tell the classifier to learn this level with more efforts. Usually, the ``focus'' of a distribution will shift from lower level to higher level (from coarse to fine). This procedure requires the classifier to extract lower features first with coarse instructions and fine tune parameters with fine instructions later. It to an extent prevents the vanishing gradient problem which would make the updates to parameters on lower layers very difficult when the network is very deep.

\section{Experiments}
\label{experiments}
\subsection{Overview}
In our experiments, first we want to see what our B-CNN is doing when handling a hierarchical classification task (in \ref{hierarchical_classification}). Then we compare the performance of B-CNN models with their corresponding baseline traditional CNN models (defined in \ref{baseline_configuration}) on benchmark datasets MNIST (in \ref{mnist}), CIFAR-10 (in \ref{cifar-10}) and CIFAR-100 (in \ref{cifar-100}). Finally we analyze our experiment results in \ref{analysis}.

In all experiments, we use stochastic gradient decent(SGD) as our optimizer with momentum set to be 0.9. All of our models are trained on Intel(R) Xeon(R) CPU E5-2637 v4, and the number of training epochs are limited to less than 80 on all datasets. This setting may result in immaturely trained models, but as we focus on the comparison between our B-CNN models and the baseline models instead of beating the current best models on these datasets, it is enough to show the idea of this paper. In our implementation, all models are trained without data augmentation and no ensemble methods are adopted.

\subsection{Hierarchical Classification}
\label{hierarchical_classification}
In this section, we choose a B-CNN model to show its behavior during training. The model is constructed with the baseline model B (second column in Table \ref{table:base}) and its corresponding branch networks (second column in Table \ref{table:branch}). The total number of training epochs is limited to 60. The learning rate is initialized with 0.003 and changes to 0.0005 at epoch 42 and 0.0001 after 52 epochs. The loss weights of the 3 branches are set to be [\emph{0.33, 0.33, 0.34}] to show their even importance. 

\begin{figure}[t]
\begin{center}
	\includegraphics[width=\linewidth]{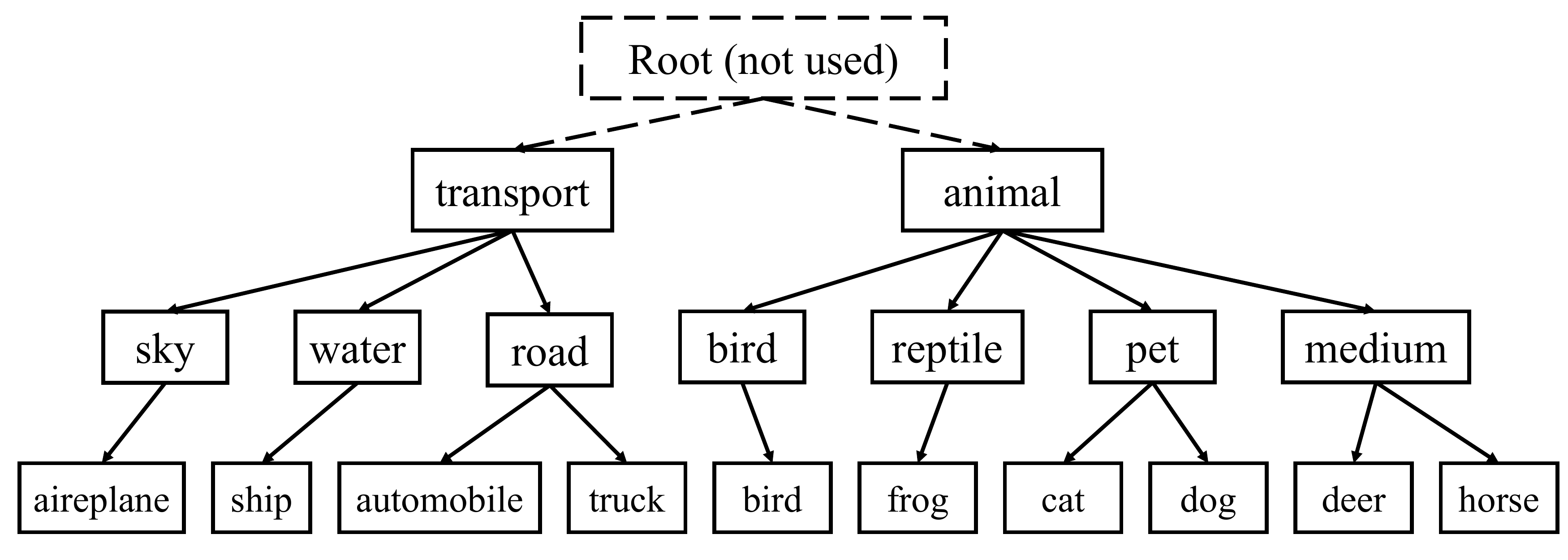}
\end{center}
\caption{A manually constructed label tree for CIFAR-10 dataset.}
\label{img:cifar-10-tree}
\end{figure}

\begin{figure}[t]
\begin{center}
	\includegraphics[width=\linewidth]{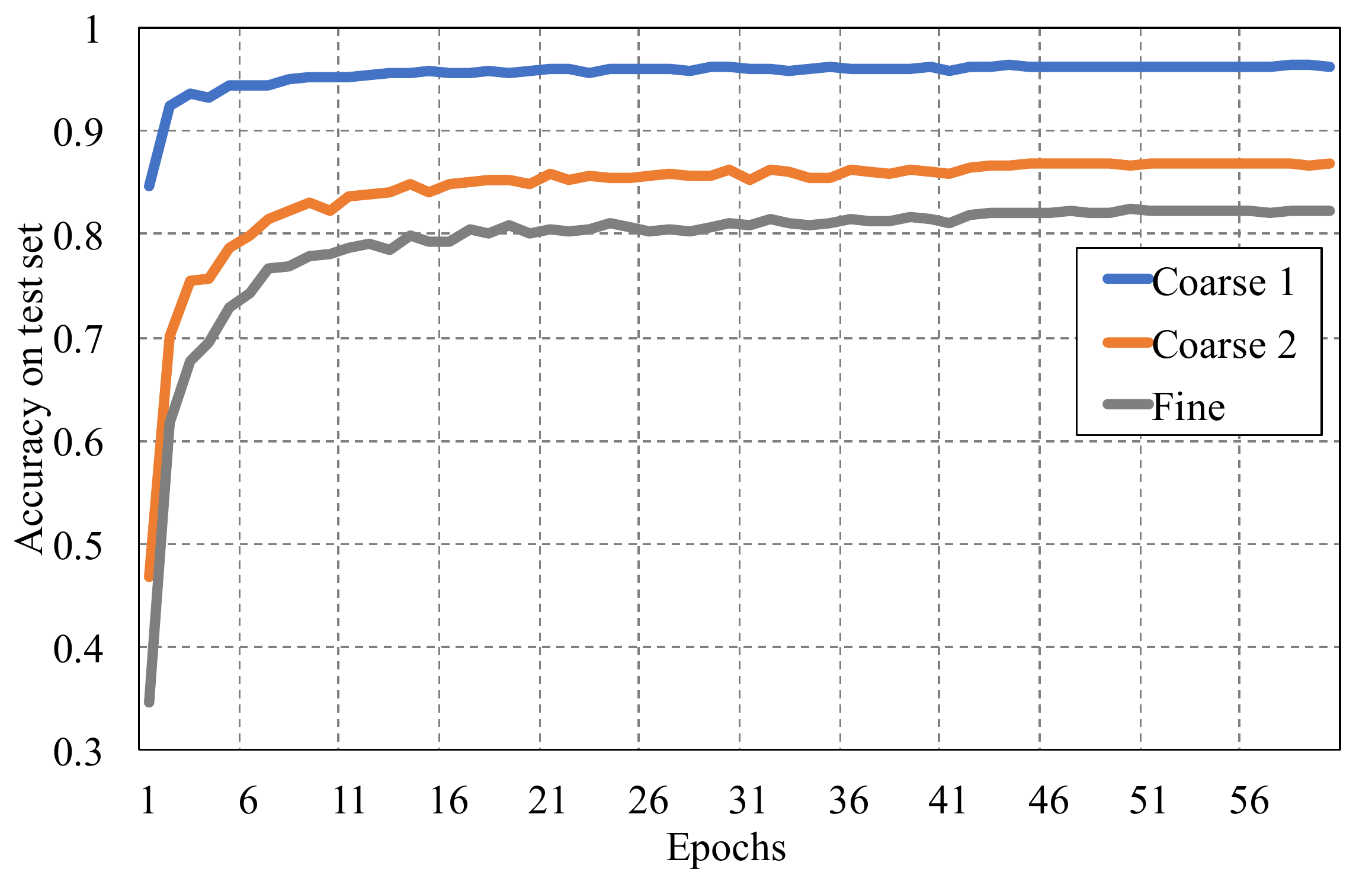}
\end{center}
\caption{Accuracy of each level based on the constructed label tree for CIFAR-10 dataset in Figure \ref{img:cifar-10-tree}.}
\label{img:hierarchy_result}
\end{figure}

The dataset used in this section is CIFAR-10. The CIFAR-10 dataset \cite{CIFAR} consists of 10 classes of $32 \times 32$ RGB images with 50,000 training and 10,000 testing examples in total. Because there is no official label tree for this dataset, we manually construct one with natural categories of the classes, \eg, \emph{cat} and \emph{dog} are clustered as \emph{pet} while \emph{deer} and \emph{horse} are grouped as \emph{medium animal}. The coarsest level consists of \emph{transport} and \emph{animal} categories. The constructed label tree is shown in Figure \ref{img:cifar-10-tree}. The model will output three accuracy results for coarse 1, coarse 2 and fine levels based on the label tree.

The results are shown in Figure \ref{img:hierarchy_result}. The coarse 1 level gets the highest accuracy and is trained fastest, followed by coarse 2 level and fine level. This is very reasonable as the coarse 1 level is the easiest one (just need to distinguish \emph{transport} and \emph{animal}) and the corresponding network is the most shallow one. On the contrary, the fine level, which is to distinguish all 10 classes, is the most difficult task among the three. Levels reach accuracy of 96.26\%, 86.74\% and 82.19\% respectively, meaning the layers before coarse 1 branch have learned enough features which can tell an image of \emph{transport} from \emph{animal} with high confidence. This confirms the usefulness of B-CNN to do a hierarchical classification and the results are consistent with our common sense.

\subsection{Baseline Configuration}
\label{baseline_configuration}
In the following sections we will compare B-CNN models with their corresponding baseline models on benchmark datasets.

The chosen baseline configurations roughly follow the construction scheme proposed by Simonyan and Zisserman's \cite{vgg}, using $3 \times 3$ patch size filters with stride fixed to 1 pixel. ReLUs are used as activation layers and max-pooling is performed over a $2 \times 2$ pixel window with stride of 2. Dropout \cite{Dropout} regularization is adopted between fully connected layers to prevent overfitting. We also use Batch Normalization \cite{BN} at each layer for easier initialization and training.

Specific configurations of three baseline models are shown in Table \ref{table:base}. Base C model is VGG16 \cite{vgg} and because this network is too deep to train from scratch without GPU, we initialize the parameters of this model with pre-trained parameters on ImageNet \cite{ILSVRC15} and fine tune it on CIFAR-10 and CIFAR-100 datasets.

The corresponding B-CNN models are just an assembling of Table \ref{table:base} and Table \ref{table:branch}. Let's take B-CNN B as an example, B-CNN B is constructed with Base B network in Table \ref{table:base} and Branches of B in Table \ref{table:branch}, where symbol * shows the conjunction. In Table \ref{table:branch}, the block [Flatten, FC-256, FC-256, FC-$c_{B1}$] is the coarse 1 branch shown in Figure \ref{img:arch} and [Flatten, FC-512, FC-512, FC-$c_{B2}$] is the coarse 2 branch. The last block of Base B in Table \ref{table:base}, [Flatten, FC-1024, FC-1024, FC-$x$], is the fine branch in Figure \ref{img:arch}, and the rest part of Base B is the ConvNet at the bottom in Figure \ref{img:arch}.

\begin{table}[t]
\begin{center}
\begin{tabular}{|c|c|c|}
\hline
Base A & Base B & Base C\\
\hline
$28 \times 28 \times 1$ image & \multicolumn{2}{|c|}{$32 \times 32 \times 3$ image}\\
\hline
conv3-32 & (conv3-64)$_{\times 2}$ & (conv3-64)$_{\times 2}$\\
\hline
maxpool-2 * & maxpool-2 & maxpool-2\\
\hline
conv3-64 & (conv3-128)$_{\times 2}$ & (conv3-128)$_{\times 2}$\\
\hhline{~--}
%\hline
& maxpool-2 * & maxpool-2\\
%\hhline{---}
\hline
conv3-64 & (conv3-256)$_{\times 2}$ & (conv3-256)$_{\times 3}$\\
\hhline{~--}
%\hline
& maxpool-2 ** & maxpool-2 * \\
\hhline{~--}
& (conv3-512)$_{\times 2}$ & (conv3-512)$_{\times 3}$\\
\hline
maxpool-2 & maxpool-2 & maxpool-2 ** \\
\hhline{~~-}
& & (conv3-512)$_{\times 3}$\\
\hline
\multicolumn{3}{|c|}{Flatten}\\
\hline
FC-128 & FC-1024 & FC-4096\\
\hline
FC-10 & FC-1024 & FC-4096\\
\hhline{~--}
& FC-$x$ & FC-$x$\\
\hline
\multicolumn{3}{|c|}{softmax layer}\\
\hline
\end{tabular}
\end{center}
\caption{Baseline networks. Base A model is tested on MNIST dataset. Both B and C are tested on CIFAR-10 and CIFAR-100. The $x$ in last ConvNet layer of Base B and C can be replaced by 10 or 100. Base C model is VGG16 \cite{vgg} without last max-pooling layer because images in CIFAR datasets are very small. Symbol * means there is a branch network attached to that layer for corresponding B-CNN models, and the specific branch configurations are shown in Table \ref{table:branch}.}
\label{table:base}
\end{table}

\begin{table}[t]
\begin{center}
\begin{tabular}{|c|c|c|}
\hline
Branches of A & Branches of B & Branches of C\\
\hline
\multicolumn{3}{|c|}{* Flatten}\\
\hline
FC-64 & FC-256 & FC-512 \\
\hline
FC-$c_{A1}$ & FC-256 & FC-512 \\
\hline
& FC-$c_{B1}$ & FC-$c_{C1}$\\
\hline
& \multicolumn{2}{|c|}{** Flatten}\\
\hline
& FC-512 & FC-1024 \\
\hline
& FC-512 & FC-1024 \\
\hline
& FC-$c_{B2}$ & FC-$c_{C2}$\\
\hline
\end{tabular}
\end{center}
\caption{Branch networks for each B-CNN model. Symbol * means this branch is connected to the layer with the same * in Table \ref{table:base}, \eg, there is a branch of network [Flatten, FC-64, FC-$c_{A1}$] after the first maxpooling layer in B-CNN A model.}
\label{table:branch}
\end{table}

\begin{figure}[t]
\begin{center}
	\includegraphics[width=0.9\linewidth]{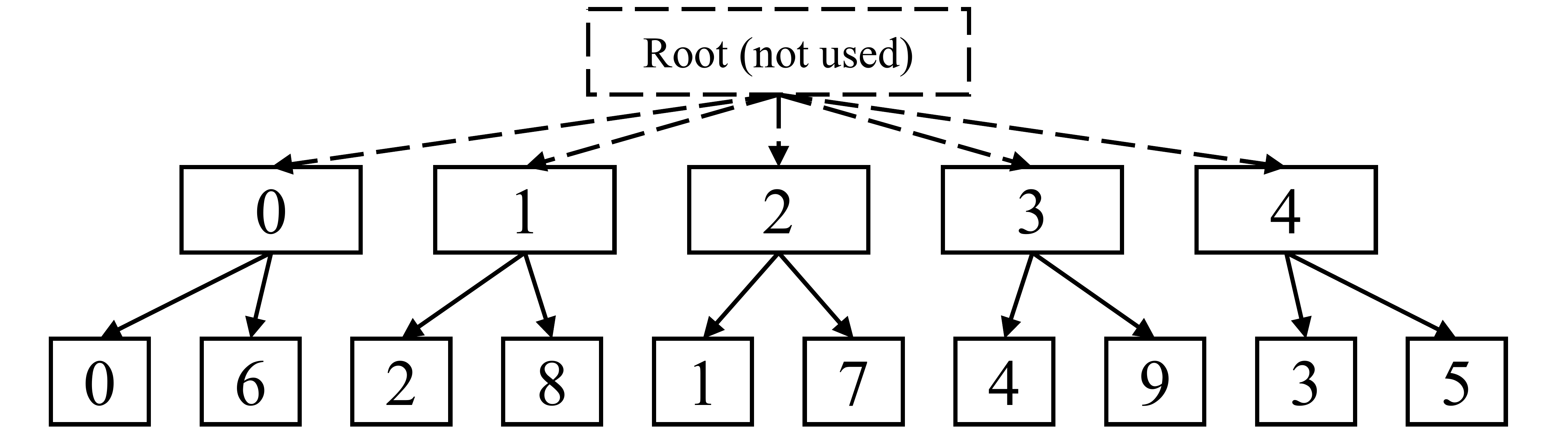}
\end{center}
\caption{A manually constructed label tree for MNIST dataset.}
\label{img:MNIST-tree}
\end{figure}

\subsection{MNIST}
\label{mnist}
The MNIST \cite{MNIST} dataset is composed of gray-scale hand written digits 0-9 which are $28 \times 28$ in size, containing 60,000 training and 10,000 testing examples.

The baseline model tested on this dataset is Base A in Table \ref{table:base}. The corresponding B-CNN model is constructed with the exact same configuration but added with a branch output network after the first maxpooling layer to do the coarse prediction (first column in Table \ref{table:branch}). Because the MNIST dataset does not provide official label tree, we manually construct one with common sense, \eg, it is reasonable to say 0 and 6 are similar. The manually constructed label tree is in Figure \ref{img:MNIST-tree}.

\begin{figure}[t]
\begin{center}
	\includegraphics[width=\linewidth]{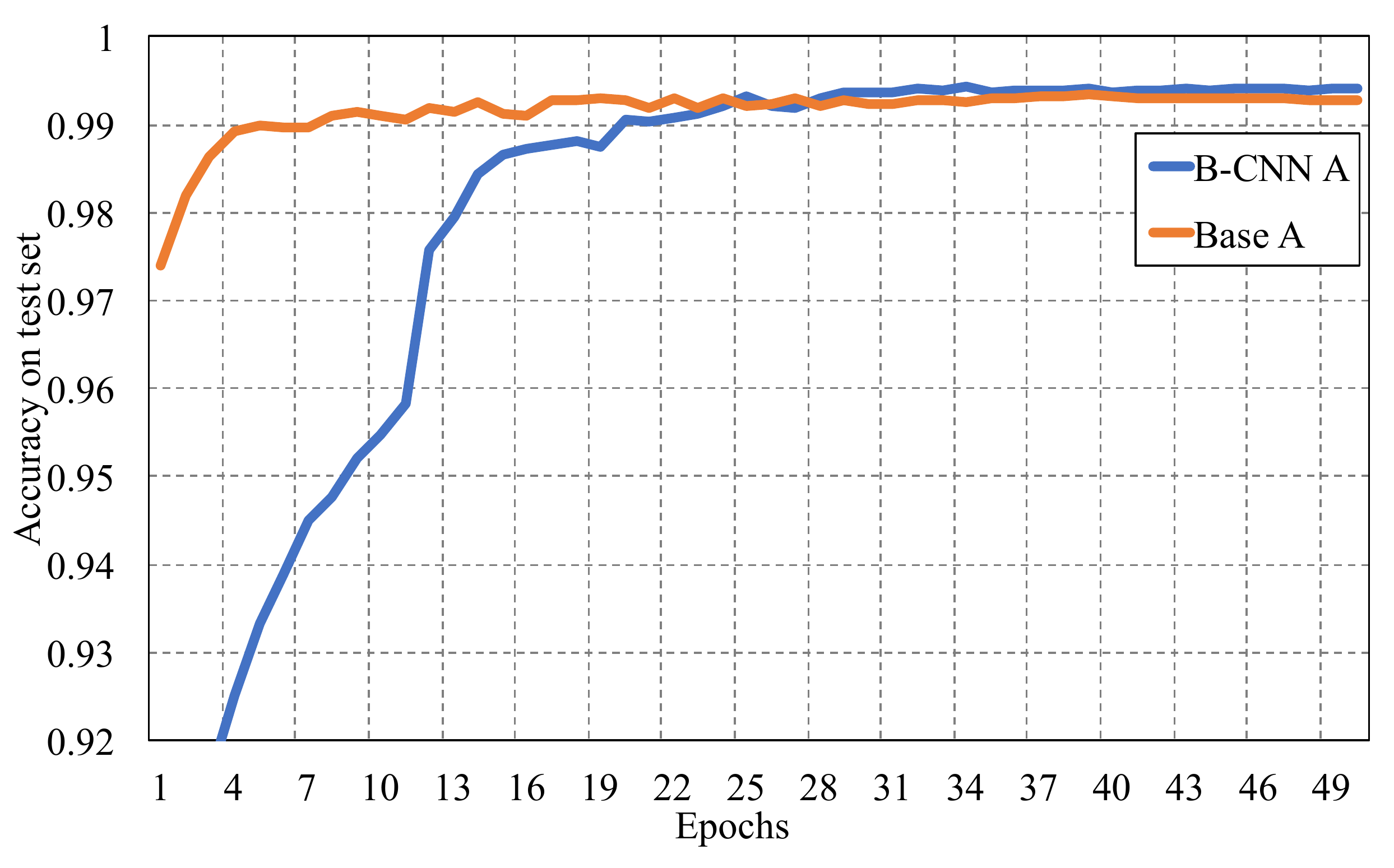}
\end{center}
\caption{Comparison on test set between baseline model A and its corresponding B-CNN model A on MNIST dataset.}
\label{img:mnist-comp}
\end{figure}

The accuracy of both models are shown in Figure \ref{img:mnist-comp}. Both models' initial learning rates are set to be 0.01 and decrease to 0.002 after 28 epochs and 0.0004 after 35 epochs. For B-CNN model, the loss weights of the two output branches will change because of the BT-strategy. Specifically, loss weights pair [$A_{1}, A_{2}$] is set to be: epoch 1: [\emph{0.98, 0.02}], epoch 12: [\emph{0.60, 0.40}], epoch 18: [\emph{0.20, 0.80}], epoch 22: [\emph{0, 1}]. When [$A_{1}, A_{2}$] changes to [\emph{0, 1}], B-CNN is converged to the traditional CNN (baseline model). As we see, the B-CNN model is trained much slower than the baseline at the beginning (baseline reaches 99\% at epoch 4 while B-CNN only gets 92\%), but it speeds up significantly at the epoch of 12. This phenomenon is caused by the change of loss weighs distribution. At the beginning, low contribution to the loss function makes the fine level prediction less important. After the loss weights change, the focus of the loss function has been shifted to the fine level and force the model to learn more specific features much faster than it did earlier.

\begin{figure}[t]
\begin{center}
	\includegraphics[width=\linewidth]{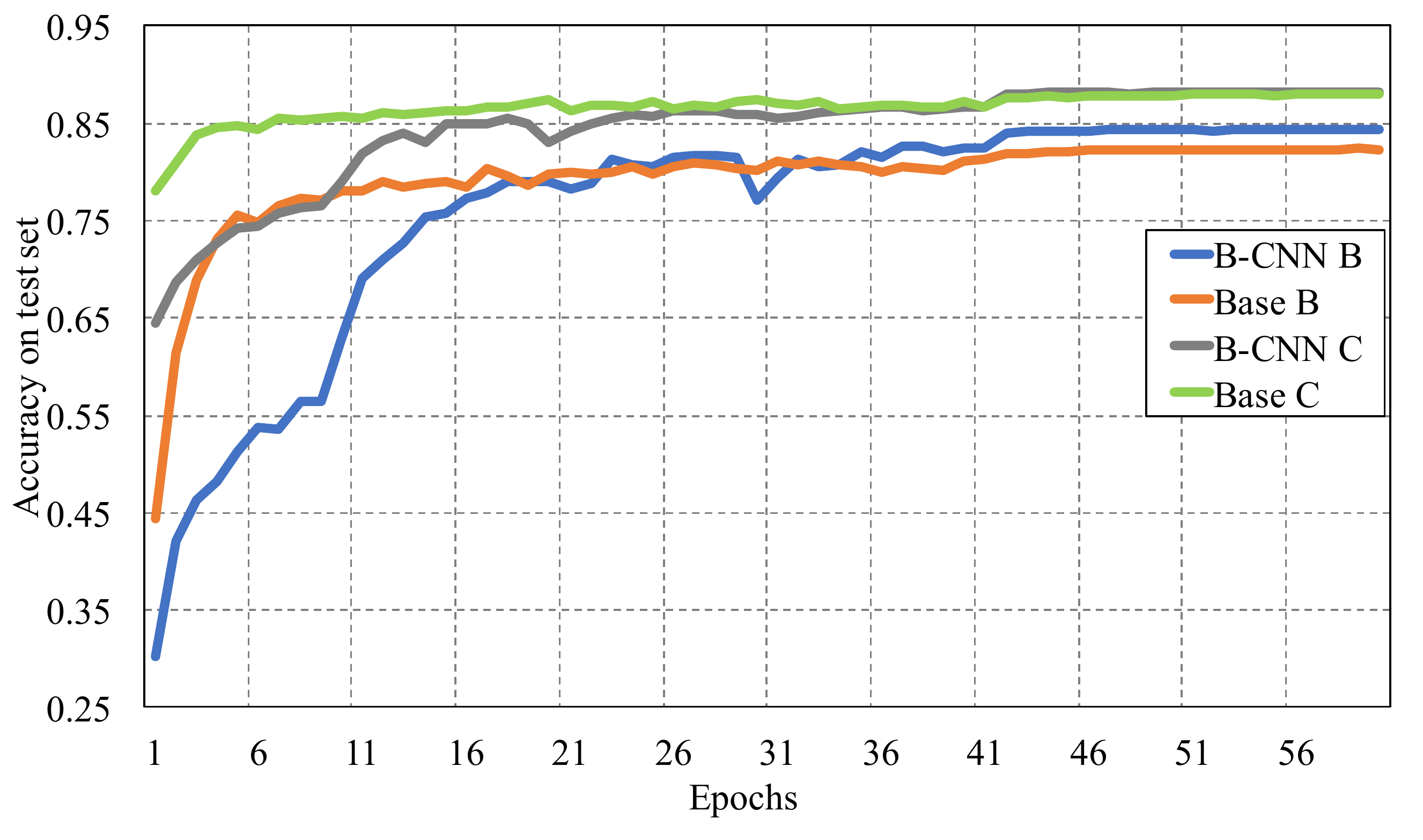}
\end{center}
\caption{Comparison on test set between baseline models and their corresponding B-CNN models on CIFAR-10 dataset.}
\label{img:cifar-10-medium-comp}
\end{figure}

\begin{figure}[t]
\begin{center}
	\includegraphics[width=\linewidth]{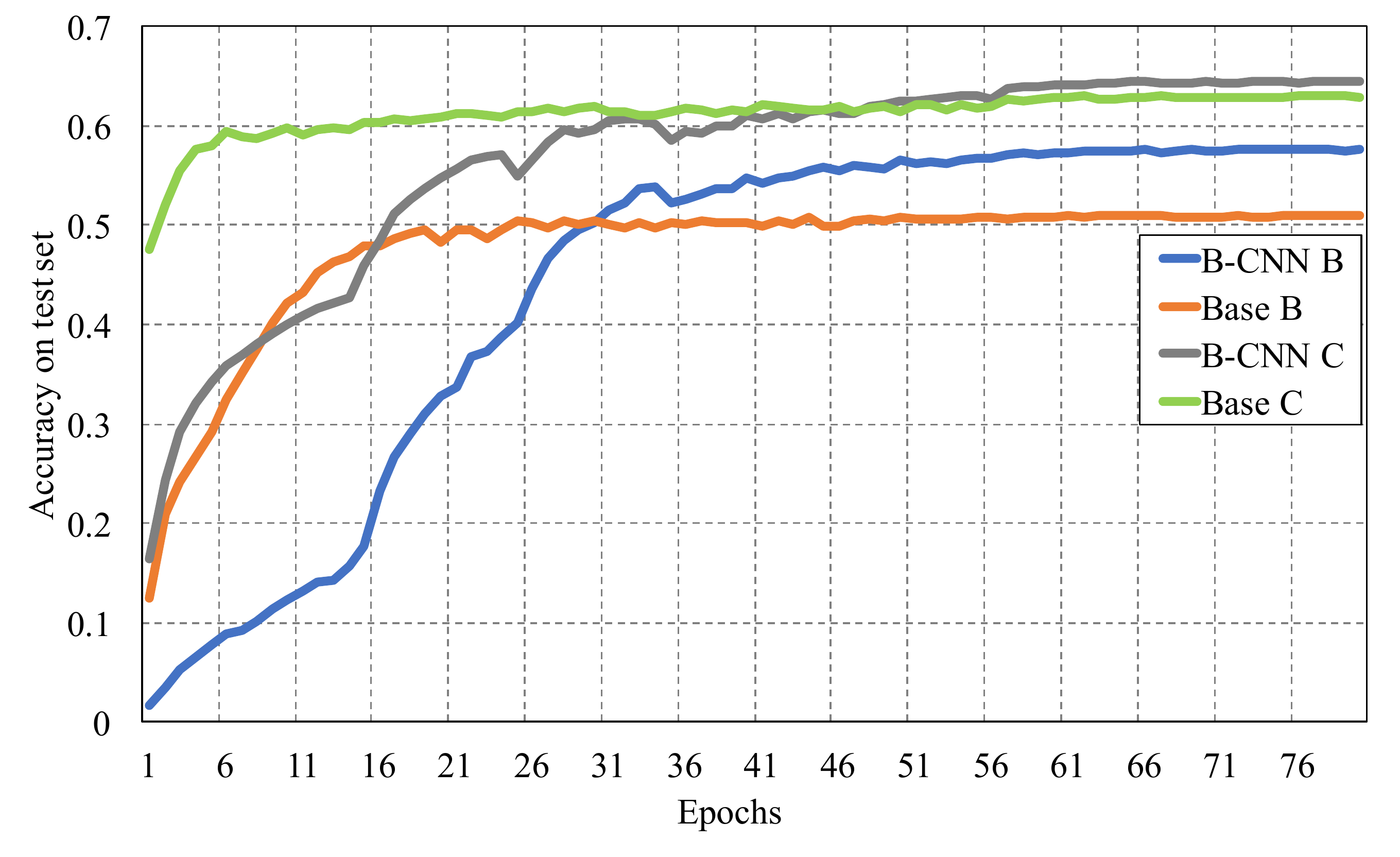}
\end{center}
\caption{Comparison on test set between baseline models and their corresponding B-CNN models on CIFAR-100 dataset.}
\label{img:cifar-100-medium-comp}
\end{figure}

\begin{table}[t]
\begin{center}
\begin{tabular}{|l|l l l|}
\hline
Models & MNIST & CIFAR-10 & CIFAR-100 \\
\hline
Base A & 99.27\% & - & - \\
B-CNN A & \textbf{99.40\%} & - & - \\
\hline
Base B & - & 82.35\% & 51.00\% \\
B-CNN B & - & \textbf{84.41\%}& \textbf{57.59\%} \\
\hline
Base C & - & 87.96\% & 62.92\% \\
B-CNN C & - & \textbf{88.22\%} & \textbf{64.42\%} \\
\hline
\end{tabular}
\end{center}
\caption{Performance of each model on MNIST, CIFAR-10 and CIFAR-100 test sets.}
\label{table:results}
\end{table}

\subsection{CIFAR-10}
\label{cifar-10}
The CIFAR-10 dataset \cite{CIFAR} consists of 10 classes of $32 \times 32$ RGB images with 50,000 training and 10,000 testing examples in total.

The baseline model tested on this dataset is Base B and Base C. Same as MNIST, the corresponding B-CNN models use the exact same configuration as baseline models with two additional branches to output extra coarse predictions (second and third columns in Table \ref{table:branch}). The corresponding label tree used in this experiment is the same one used in \ref{hierarchical_classification} (shown in Figure \ref{img:cifar-10-tree}).

For model B, both models' learning rates are initialized to be 0.003 and decease to 0.0005 after 42 epochs and 0.0001 after 52 epochs. For B-CNN model, the change of loss weights follows the scheme: epoch 1: [\emph{0.98, 0.01, 0.01}], epoch 10: [\emph{0.10, 0.80, 0.10}], epoch 20: [\emph{0.1, 0.2, 0.7}], epoch 30: [\emph{0, 0, 1}].

For model C, we use the pre-trained parameters of VGG16 on ImageNet dataset to initialize both the baseline and B-CNN models then fine tune them on CIFAR-10 dataset, using the same training procedure for model B.

The result is shown in Table \ref{table:results} and Figure \ref{img:cifar-10-medium-comp}. The baseline model B gets an accuracy of 82.35\% after 60 epochs and the B-CNN model B reaches 84.41\%. For B-CNN model B, the loss weight of the fine-level loss increases at epoch 10, so there is an obvious tendency shift at that point. As we see, the new growth trend is much steeper than the old one, which indicates that the network is putting more effort to learn more specific features. For baseline C and B-CNN C, the training procedure is very similar to models of B, but the final accuracy gap between B-CNN and baseline is smaller. We presume the different initialization caused this difference as for B the parameters are initialized randomly while for C they are initialized with pre-trained parameters, which makes the BT-strategy less important.

\subsection{CIFAR-100}
\label{cifar-100}
The CIFAR-100 dataset \cite{CIFAR} is just like CIFAR-10 with $32 \times 32$ images, except that it has 100 classes containing 600 images each. The 100 fine classes in the CIFAR-100 are grouped into 20 coarse classes. In this case, we also manually group the provided 20 superclasses into 8 coarser classes as a more informative prior. The corresponding B-CNN models are with same configuration but extra branches shown in second and third columns in Table \ref{table:branch}.

The training procedures on CIFAR-100 are very similar to the ones on CIFAR-10 except the learning rate and loss weights which are determined by cross validation. The learning rates of all models trained on CIFAR-100 are initialized as 0.001 and drop to 0.0002 at epoch 55 and 0.00005 after epoch 70. The loss weights modification scheme for B-CNNs are: epoch 1: [\emph{0.98, 0.01, 0.01}], epoch 15: [\emph{0.10, 0.80, 0.10}], epoch 25: [\emph{0.1, 0.2, 0.7}], epoch 35: [\emph{0, 0, 1}]. Same as on CIFAR-10, the type C models are initialized with existent VGG16 parameters pre-trained on ImageNet dataset.

The performance of models trained on CIFAR-100 are shown in Table \ref{table:results} and Figure \ref{img:cifar-100-medium-comp}. For model B, the baseline model gets 51.00\% accuracy while its corresponding B-CNN model reaches 57.59\%. Though this model may not be fully trained as we limit the number of epochs to 80, it can be seen in Figure \ref{img:cifar-100-medium-comp} that B-CNN model can reach high accuracy faster than the baseline model. The baseline model C and B-CNN model C get 62.92\% and 64.42\% accuracy respectively. Note that same as in CIFAR-10, the accuracy gap between baseline and B-CNN on model C is smaller than that on model B. This may be caused by the initialization of pre-trained parameters, which makes it much easier for the baseline model to fine tune weights.

\subsection{Analysis}
\label{analysis}
These experiments on three different datasets share many features. First the learning pattern of B-CNNs differs from traditional CNNs. There is an obvious acceleration of learning speed after the loss weights contribution shift in B-CNN models. This phenomenon confirms that using coarse level labels to learn low level features first is very useful to activate the shallow layers of a CNN model. In other words, our BT-strategy prevents B-CNNs from suffering the vanishing gradient problem. Second, B-CNN models consistently outperform their corresponding baseline models. This result is a convincing evidence that hierarchical nature of CNNs can be connected to the structure prior of target classes to strengthen the classifier. Third, B-CNN models are as simple as traditional CNNs. B-CNNs only use existent CNN components as building blocks and BT-strategy is achieved as easily as modifying learning rates. Forth, a good initialization may downgrade the benefit of B-CNN. When models are initialized with pre-trained parameters, the performance gap between B-CNN and baseline models is not that obvious. We presume it is because when pre-trained parameters are used, the low level features have already been successfully extracted so the benefit of BT-strategy is not very obvious.

\section{Conclusion}
\label{conclusion}

%In this paper, we introduce Branch Convolutional Neural Network (B-CNN) which connects the hierarchical nature of CNNs and structure prior of target classes. Compared with traditional CNN models, B-CNN can output multiple hierarchical predictions from coarse to fine, which is more informative and interpretable. We also introduce B-CNN's tailored Branch Training strategy (BT-strategy) to enforce the model to learn low level features at the beginning and converge to traditional CNN classification at last. This strategy enables B-CNN models to utilize label trees as internal guides and boosts the performance significantly. The experiment results confirm the benefits of our model over the traditional CNN.

In this paper, we introduced the Branch Convolutional Neural Network
(B-CNN) which connects the hierarchical nature of CNNs and a structured
prior on target classes. Compared with traditional CNN models, B-CNN can
output multiple hierarchical predictions from coarse to fine, which is
more informative and interpretable. We also introduce B-CNN's tailored
Branch Training strategy (BT-strategy) to force the model to learn
low-level features at the beginning of training and converge later to
traditional CNN classification. This strategy enables B-CNN models to
utilize label trees as internal guides, and boosts performance significantly.
The experiment results confirm the benefits of our model over the
traditional CNN. For further work the possibility of learning with other
structured outputs, such as linear chains or graphs, should be
investigated.

{\small
\bibliographystyle{ieee}
\bibliography{./ref}
}

\end{document}